\documentclass{article}

\usepackage[preprint,nonatbib]{neurips_2024}

\usepackage[utf8]{inputenc} 
\usepackage[T1]{fontenc}    
\usepackage{hyperref}       

\usepackage{url}            
\usepackage{booktabs}       
\usepackage{amsfonts}       
\usepackage{nicefrac}       
\usepackage{microtype}      
\usepackage{xcolor}         
\usepackage{algorithm}
\usepackage{algorithmic}
\usepackage{amssymb}
\usepackage{multirow}
\usepackage{graphicx}
\usepackage{subfigure}
\usepackage{array, boldline}
\usepackage{dashrule}
\usepackage{amsmath}
\usepackage{mathtools}
\usepackage{amsthm}
\usepackage{tabularx}
\usepackage{bbm}
\usepackage{wrapfig}
\usepackage{enumitem}
\usepackage{caption}
\usepackage{nicematrix}
\usepackage{color, colortbl}
\usepackage[capitalize,noabbrev]{cleveref}

\usepackage{amsmath,amsfonts,bm}

\def\eqref#1{equation~\ref{#1}}

\def\1{\bm{1}}

\def\mM{{\bm{M}}}

\DeclareMathAlphabet{\mathsfit}{\encodingdefault}{\sfdefault}{m}{sl}
\SetMathAlphabet{\mathsfit}{bold}{\encodingdefault}{\sfdefault}{bx}{n}

\usepackage{wrapfig}

\makeatletter
\renewcommand\@cite[2]{[{#1\if@tempswa , #2\fi}]}
\renewcommand\@biblabel[1]{[{#1}]}
\makeatother

\newcommand{\gr}{\rowcolor[gray]{.95}}
\newcommand{\cellr}{\cellcolor[gray]{.95}}

\definecolor{lightgray}{rgb}{0.7,0.7,0.7}
\definecolor{myred}{rgb}{0.8535, 0.07, 0.07}
\newcolumntype{P}[1]{>{\centering\arraybackslash}p{#1}}

\title{Enhancing One-shot Pruned Pre-trained Language Models through Sparse-Dense-Sparse Mechanism}

\author{%
  Guanchen Li\footnotemark[1]\textsuperscript{,}\footnotemark[2]
 \\
  \texttt{guanchen.li@amd.com} \\
  \And
  Xiandong Zhao\footnotemark[1]\textsuperscript{,}\footnotemark[2] \\
  \texttt{xiandong.zhao@amd.com} \\
  \And
  Lian Liu\footnotemark[2] \\
  \texttt{lian.liu@amd.com} \\
  \And
  Zeping Li\footnotemark[2] \\
  \texttt{zeping.li@amd.com} \\
  \And
  Dong Li\footnotemark[2] \\
  \texttt{d.li@amd.com} \\
  \And
  Lu Tian\footnotemark[2] \\
  \texttt{lu.tian@amd.com} \\
  \And
  Jie He\footnotemark[3] \\
  \texttt{hejie@ustb.com} \\
  \And
  Ashish Sirasao\footnotemark[2] \\
  \texttt{ashish.sirasao@amd.com} \\
  \And
  Emad Barsoum\footnotemark[2] \\
  \texttt{emad.barsoum@amd.com} \\
}

\begin{document}

\maketitle
\footnotetext[1]{Equal contribution.}
\footnotetext[2]{Advanced Micro Devices, Inc.}
\footnotetext[3]{The School of Computer and Communication Engineering, University of Science and Technology Beijing.}

\ifx\allfiles\undefined

\fi
\begin{abstract}
Pre-trained language models (PLMs) are engineered to be robust in contextual understanding and exhibit outstanding performance in various natural language processing tasks. However, their considerable size incurs significant computational and storage costs.
Modern pruning strategies employ one-shot techniques to compress PLMs without the need for retraining on task-specific or otherwise general data; however, these approaches often lead to an indispensable reduction in performance.
In this paper, we propose \textbf{SDS}, a \textbf{S}parse-\textbf{D}ense-\textbf{S}parse pruning framework to enhance the performance of the pruned PLMs from a weight distribution optimization perspective. We outline the pruning process in three steps. Initially, we prune less critical connections in the model using conventional one-shot pruning methods. Next, we reconstruct a dense model featuring a pruning-friendly weight distribution by reactivating pruned connections with \textit{sparse regularization}. Finally, we perform a second pruning round, yielding a superior pruned model compared to the initial pruning.
Experimental results demonstrate that SDS outperforms the state-of-the-art pruning techniques SparseGPT and Wanda under an identical sparsity configuration. For instance, SDS reduces perplexity by 9.13 on Raw-Wikitext2 and improves accuracy by an average of 2.05\% across multiple zero-shot benchmarks for OPT-125M with 2:4 sparsity.

\end{abstract}

\ifx\allfiles\undefined
\fi
\ifx\allfiles\undefined

\fi
\section{Introduction}\label{sec:introduction}

Pre-trained language models (PLMs) \cite{attention} have revolutionized various applications in natural language processing. However, the considerable size of PLMs results in notable drawbacks, such as increased latency and energy consumption. Compression methods for vision models such as convolutional neural networks, which perform \textit{pre-training, compression, and fine-tuning} workflow with quantization or pruning \cite{cnn_compression}, may be ill-suited for PLMs due to their prohibitive training cost.

Recent pruning research, such as SparseGPT \cite{sparsegpt} and Wanda \cite{wanda}, has introduced effective one-shot compression techniques for PLMs. These methods can compress up to 50\% of the parameters in the fully connected layers of ultra-large PLMs with negligible impact on performance. However, their effectiveness diminishes when applied to compact ones, which are usually more thoroughly trained. For instance, SparseGPT, the state-of-the-art pruning method, yields a perplexity of 31.58 when applied to prune 50\% of the weights in OPT-350M. This score is worse than the 27.66 perplexity observed in OPT-125M, a dense model with roughly half the parameters of OPT-350M. Furthermore, when stricter sparsity constraints are employed, such as 2:4 or 4:8 sparse configurations \cite{nmsparse} for computational acceleration, the performance deteriorates even further. Therefore, it is essential to optimize the poorly pruned PLMs.

Compact PLMs cover not only undersized ones but also more fully trained models. On the one hand, compact PLMs are less over-parameterized and naturally harder to compress. On the other hand, PLMs are not designed to be aware of subsequent pruning since they lack pruning-related regularization during pre-training. As a result, pruning compact PLMs while maintaining their performance proves challenging.

\begin{wraptable}{r}{6.2cm}
\vspace{-1em}
    \caption{\textbf{Pruning PLMs Incurs Resumable Knowledge Loss.} We apply 2:4 sparse to OPTs with SparseGPT, and their performance decreases on Raw-WikiText2. However, a substantial performance improvement is observed upon reactivating the sparse weights with only 128 samples from C4.}
    \label{tab:no_loss}
    \begin{center}
    \renewcommand{\arraystretch}{1.2} 
    \resizebox{6.2cm}{!}{
    \begin{tabular}{cccc}
    \hlineB{3}
    PLMs & Dense & 2:4 Sparse & Re-dense \\ \hline
    OPT-125M & 27.66 & 60.43 & 27.94 \\
    OPT-350M & 22.01 & 51.11 & 22.25 \\ 
    \hlineB{3}
\end{tabular}}
    \end{center}
\end{wraptable}


Neurons in the human brain show sparse-to-dense-to-sparse connectivity as they grow \cite{connectivity}. This observation inspired us to perform a similar process to achieve a better pruning scheme that benefits from pruning friendliness.
We preliminarily explored layer-wise dense reconstruction to find a performance upper bound. Intriguingly, we discovered that sparse models could bounce back to performance levels equivalent to their dense counterparts using only a few samples (cf., \Cref{tab:no_loss}). It reveals two key insights: first, pruned PLMs can be optimized with limited resources; second, the amount of knowledge lost during the pruning process is restorable. These insights lay the foundation for the work presented in this paper.

We propose a three-step \textbf{S}parse-\textbf{D}ense-\textbf{S}parse (\textbf{SDS}) pruning framework to enhance the performance of pruned pre-trained language models.
\textbf{In the first step}, we employ conventional one-shot pruning methods on a PLM to remove irrelevant connections. 
\textbf{In the second step}, we perform a dense reconstruction of the sparse model to reactivate the pruned connections, aiming to identify a dense model with enhanced pruning awareness. This process is aided by a multidimensional sparse regularization strategy, which optimally guides the weight distribution, rendering it more pruning-friendly for the subsequent step.
\textbf{In the third step}, we further prune and adjust the weights of the second-pruned model. 
Importantly, SDS requires only a limited number of samples for calibration, identical to conventional one-shot methods. 
Experimental results demonstrate that SDS outperforms SparseGPT and Wanda under the same sparsity configuration. For example, SDS reduces perplexity by 9.13 on Raw-Wikitext2 and increases average accuracy by 2.05\% across multiple downstream tasks for OPT-125M with 2:4 sparsity. 
The pruned PLMs achieve up to 1.87x acceleration on an AMD R7 Pro CPU.
The main contributions of the paper are summarized as follows:

\begin{itemize}[itemsep=0mm, leftmargin=*]
    \item We introduce SDS, a three-step Sparse-Dense-Sparse framework. It involves weight redistribution and pruning, enhancing the performance of the one-shot pruned pre-trained language models.
    \item We design sparse regularization strategies that improve the effectiveness of re-dense weight reconstruction and find a more pruning-friendly weight distribution.
    \item Experimental results demonstrate that SDS outperforms existing pruning methods in language comprehension and downstream task performance.
\end{itemize}

\ifx\allfiles\undefined
\fi
\ifx\allfiles\undefined

\fi
\section{Sparse-Dense-Sparse Framework}\label{sec:method}

\begin{figure*}[t]
\centering
\includegraphics[width=\linewidth]{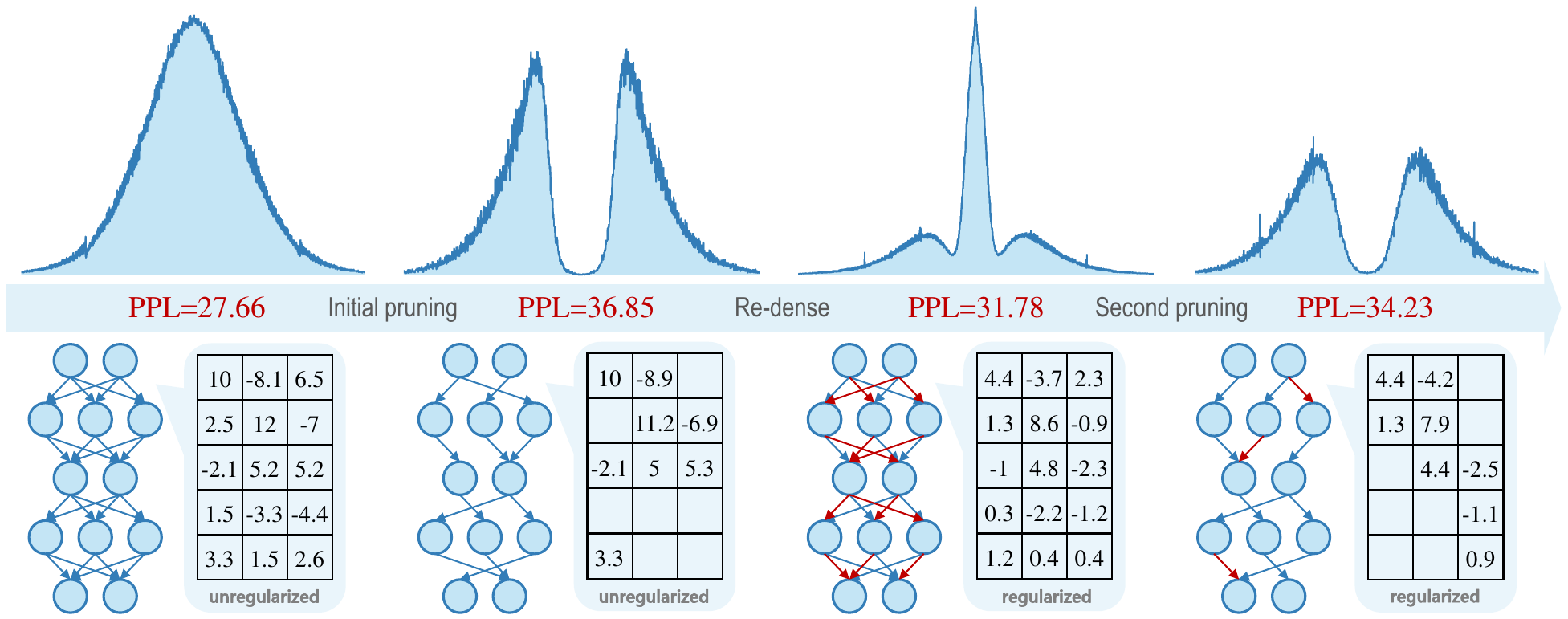}
\caption{\textbf{An Overview of the Steps of the SDS Framework}, divided into initial pruning, re-dense weight reconstruction, and a second round of pruning. The upper figure shows the weight distribution variation within the SDS framework, and the lower figure demonstrates the variation in weight connections. The weights are extracted from the FFN in the $12$-th transformer block of OPT-125M, with 50\% sparsity configuration. Initially, the dense weights follow a Gaussian distribution. After being pruned by SparseGPT, a concentrated, bimodal distribution emerges (zero values are omitted in sparse weight distributions for better clarity). Followed by connection reconstruction with sparse regularization, a three-peaked distribution materializes. Finally, the second pruning round attenuates the sharp peaks, resulting in a softer bimodal distribution. Perplexity (PPL) is evaluated on Raw-WikiText2. The second pruned model achieves a lower perplexity than the initially pruned one.}
\label{fig:overview}
\end{figure*}

This section presents the Sparse-Dense-Sparse (SDS) framework to perform optimization for pruned pre-trained language models (PLMs). Firstly, we provide a brief overview of the core Transformer architecture, which is fundamental to most PLMs. A standard Transformer block consists of two main modules: a multi-head attention (MHA) layer and a feed-forward network (FFN). Let $\mathbf{X}_{\ell-1} \in \mathbb{R}^{d \times n}$ represent the input of the $\ell$-th Transformer block, where $n$ is the sequence length, and $d$ is the size of the hidden state. The block output $\mathbf{X}_{\ell}$ can be formulated as:
\begin{equation}
\label{eq:1}
\begin{aligned}
    \mathbf{X} &= \operatorname{MHA}\left(\text{ LayerNorm }\left(\mathbf{X}_{\ell-1}\right)\right)+\mathbf{X}_{\ell-1}, \ \ \ \mathbf{X}_\ell &= \mathrm{FFN}(\text { LayerNorm }(\mathbf{X}))+\mathbf{X}.
\end{aligned}
\end{equation}
MHA consists of $h$ heads, represented as $\mathbf{W}^{\text{O}}\cdot \texttt{concat}(\text{head}_1, \text{head}_2, \ldots, \text{head}_h) $, with $\mathbf{W}^\text{O}$ responsible for the output projection. Specifically, the $i$-th head can be expressed as:
\begin{gather}
    \text{head}_i = \text{Attn}([\mathbf{W}^{\text{Q}}\mathbf{X}]_i, [\mathbf{W}^{\text{K}}\mathbf{X}]_i, [\mathbf{W}^{\text{V}}\mathbf{X}]_i, \mM), \\
    \text{Attn}(\mathbf{Q}, \mathbf{K}, \mathbf{V}, \mathbf{M}) = \text{softmax}\left( \mathbf{M} \odot \frac{\mathbf{Q} \mathbf{K}^\top}{\sqrt{d_\text{K}}}\right) \mathbf{V},
\end{gather}
where $\mathbf{Q}$, $\mathbf{K}$, and $\mathbf{V}$ represent the query, key, and value sequences, respectively, and their corresponding projection weights are $\mathbf{W}^\text{Q}$, $\mathbf{W}^\text{K}$, and $\mathbf{W}^\text{V}$. $d_\text{K}$ is the dimension of the key vectors, and $\mathbf{M}$ is the mask matrix to selectively ignore or give weight to specific tokens in the input sequence. FFN expands and contracts input dimensions through hidden layers, introducing non-linearities to enhance representation learning, which consists of several fully connected layers, with their weights represented as  $\mathbf{W}^\text{Up}$, $\mathbf{W}^\text{Down}$, and $\mathbf{W}^\text{Gate}$ (optional) respectively.

In this paper, we focus on pruning the weights in fully connected layers $\mathbf{W}$, which are emphasized by $\mathbf{W}^\text{Q}$, $\mathbf{W}^\text{K}$, $\mathbf{W}^\text{V}$, $\mathbf{W}^\text{O}$, $\mathbf{W}^\text{Up}$, $\mathbf{W}^\text{Down}$ and $\mathbf{W}^\text{Gate}$ from the outset.

The SDS framework consists of three steps: initial pruning (\Cref{sec:2-1}), re-dense weight reconstruction (\Cref{sec:2-2}), and a second round of pruning (\Cref{sec:2-3}). By optimizing weight distribution through these steps, the SDS framework enhances the performance of pruned PLMs. The SDS framework's overall process and weight distribution evolution are depicted in \Cref{fig:overview}.

\subsection{Initial Pruning}
\label{sec:2-1}

\begin{wraptable}{r}{7.8cm} \small
\vspace{-1.3em}
\caption{\textbf{Saliency Metric and Weight Update Rules for Conventional Pruning Method}.}\label{table:metric}
\begin{center} 
\renewcommand{\arraystretch}{1.2}
\begin{tabular}{ccc}
\hlineB{2.5}
Method    & Salience Metric                                                        & Weight Update $\mathbf{\Delta}$                                                                                                             \\ \hline
SparseGPT & $\frac{{{\mathbf{W}}_{:,c}^{\text{dense}}}^2}{[\mathbf{H}^{-1}]^2_{c,c}}$ & $\frac{\mathbf{W}^{\text{dense}}_{:,c} - \mathbf{W}_{:,c}^{\text{sparse}}}{[\mathbf{H}^{-1}]_{c,c}}  \cdot [\mathbf{H}^{-1}]_{c, c:}$ \\
Wanda     & $|\mathbf{W}^{\text{dense}}| \odot\|\mathbf{X}\|_2$     & - {\tiny Update Free} -                                                                                                                                  \\ \hlineB{2.5}
\end{tabular}
\end{center}
\end{wraptable}

The SDS framework initiates by eliminating the less critical connections in PLMs using conventional one-shot pruning methods.
SparseGPT \cite{sparsegpt} leverages second-order information to guide sparse mask selection and update weights. Concretely, during column-wise pruning, SparseGPT compensates for the pruning error of the pruned columns (prior to column $c$) by updating the unpruned columns (subsequent to column $c$) of the weight matrix ($\mathbf{W}_{:, c:} = \mathbf{W}_{:, c:} - \mathbf{\Delta}$). Wanda \cite{wanda} switches the perspective to pruning mask selection and realizes weight update-free pruning by considering both weight and activation magnitude. The salience metrics and weight updates for the two preferred pruning methods are shown in \Cref{table:metric}.

SparseGPT and Wanda demonstrate robust performance on ultra-large models such as 70B and 175B, achieving negligible performance degradation. However, its efficacy diminishes when applied to compact ones. Firstly, the parameters within compact models are more uniformly de-tended to be adequately expressed and are more challenging to compress. Secondly, the weight distribution of the original dense models is inappropriate for direct pruning due to the lack of sparse regularization during pretraining. Thus, we take SparseGPT / Wanda as the initial step and identify a superior sparse model from the perspective of weight distribution optimization in the subsequent steps.

\subsection{Re-dense Weight Reconstruction}
\label{sec:2-2}

\Cref{tab:no_loss} demonstrated that there is more than one possible set of dense-weight solutions with similar performance. To this end, our goal is to find a weight solution that exhibits pruning awareness and forms \textbf{a pruning-friendly dense model} to serve as a new starting point for pruning. 
Concretely, we implement layer-wise knowledge distillation with the same samples as the initial pruning step to reactivate the connections in pruned PLMs. This method guarantees high efficiency while preserving the unbiased capabilities of the re-dense PLMs.

However, a na\"ive re-dense weight reconstruction is insufficient. We introduce three sparse regularization strategies to circumvent ending up with a re-dense solution resembling the original dense model.
\textbf{a) Residual sparse characteristics}: the initial pruning cannot be omitted; it provides prior information about which weights are relatively important for the re-dense weight reconstruction process.
\textbf{b) Data-based regularization}: hard-to-learn samples (high-loss sparse-model activations) are used as the inputs of re-dense weight reconstruction to avoid overfitting. 
\textbf{c) Weight-based regularization}: typical weight regularization is also employed to endow the re-dense weights with sparse features, thereby directly increasing the pruning friendliness.
We choose the L1 and L2 regularization \cite{l1_reg, weight_decay} to meet the requirement.


According to the above considerations, the re-dense weight reconstruction process is specified in the following. Given the original dense weight $\mathbf{W}^{\text{dense}}_{\ell}$ in layer $\ell$, the sparse weight $\mathbf{W}^{\text{sparse}}_{\ell}$ from the initial pruning step, and $\mathbf{X}_{\ell-1}$ collected during the forward propagation of the sparse model, the re-dense weight $\widehat{\mathbf{W}}_{\ell}^{\text{re-dense}}$ is obtained by:


\begin{equation}
\label{eq:3}
\begin{aligned}
\widehat{{\mathbf{W}}}_{\ell}^{\text{re-dense}} = \operatorname{argmin}_{{\mathbf{W}}^{\text{sparse}}_{\ell}} \bigg( &\left\| {\mathbf{W}}^{\text{dense}}_{\ell} {\mathbf{X}}_{\ell-1} - {\mathbf{W}}^{\text{sparse}}_{\ell} {\mathbf{X}}_{\ell-1} \right\|_2^2
+\lambda_1 \|{\mathbf{W}}^{\text{sparse}}_{\ell}\|_1 + \lambda_2 \|{\mathbf{W}}^{\text{sparse}}_{\ell}\|_2 \bigg),
\end{aligned}
\end{equation}

where $\lambda_1$ and $\lambda_2$ are used to control the ratio of the L1 and L2 regularization, they are set to be 0.1 by default. The distribution of $\widehat{\mathbf{W}}_{\ell}^{\text{re-dense}}$ are shown in \Cref{fig:overview}. 
Firstly, the parameters obtained through re-dense weight reconstruction show a clear three-peaked distribution. This distribution displays a higher sharpness around zero than the original dense model, which is a terrific phenomenon. It indicates that the increased concentration of values around zero makes irrelevant weights easier to identify, a trait referred to as pruning-friendliness \cite{dsd}.
Secondly, the re-dense weight reconstruction yields a model with performance slightly below that of the original dense model but significantly better than the initial sparse model, aligning with our expectations: under the constraints of regularization, it is straightforward that the performance of the re-dense model struggles to maintain consistency with the original dense model. \Cref{app:2} provides a more detailed analysis for sparse regularization. 

\subsection{Second Pruning: Sparse Weight Adjustment}
\label{sec:2-3}


Directly adjusting weights in a Sparse-to-Sparse manner seems intuitive for enhancing a sparse model's performance; however, when applied to a model after the initial pruning stage, it only results in minor performance gains on the lightest models (cf., \Cref{table:abl}). This approach simply introduces a first-order loss term to guide the layer-wise optimization, which is insufficient.

Considering the aforementioned challenges, we introduce sparse weight adjusting as the concluding step in the SDS framework. The re-dense model obtained with sparse regularization guidance will inevitably perform inferior to the pre-trained model. As a result, directly pruning it might not be ideal. Therefore, we perform weight adjustment for the second-pruned model. To elaborate, we first prune the re-dense model using the same method employed during the initial pruning, yielding $\mathbf{W}^{\text{sparse-2nd}}$. Subsequently, weight adjusting is conducted utilizing a soft sparse mask:


\begin{equation}
\label{eq:4}
\begin{aligned}
{\widehat{\mathbf{W}}}_{\ell}^{\text{SDS}} = 
{\textbf{Mask}}^{\text{soft}}_{\ell} \odot
\bigg(\operatorname{argmin}_{\mathbf{W}^{\text{sparse-2nd}}_{\ell}} \left\|\mathbf{W}^{\text{dense}}_{\ell} \mathbf{X}_{\ell-1} 
-\mathbf{W}^{\text{sparse-2nd}}_{\ell}\mathbf{X}_{\ell-1}\right\|_2^2\bigg),
\end{aligned}
\end{equation}

where $\mathbf{X}_{\ell-1}$ is collected from the forward propagation of the second pruned model. ${\textbf{Mask}}^{\text{soft}}_{\ell}$ represents a soft sparse mask, which is dynamically selected by $|\mathbf{W}_\ell^{\text{sparse-2nd}}|$ in each iteration. Due to the inherent awareness of activation information from backpropagation, this magnitude (absmin) \cite{magnitude} mask selection metric can achieve results similar to the elaborate salience metric in SparseGPT and Wanda. In both steps of weight adjustment, the L2 loss is utilized, inherently emphasizing the loss in regions with outliers \cite{smoothq}, which plays a pivotal role in the performance of language models. Therefore, outliers can be protected and less affected by weight adjustments. Notably, for most models above three billion parameters, it is possible to do direct one-shot pruning of the re-dense model and exceed the performance of initial pruning. At this point, we choose to \textit{early-exit}, skipping the weight adjustment process and, therefore, more efficient.

As shown in \Cref{fig:overview}, the weights presented after the second pruning became moderate, i.e., the weight distribution of the second-pruned model is smoother and more uniform than that of the initial pruning step, which means that the model parameters have suitable values in different ranges, possessing a stronger ability to adapt to unseen data.

\ifx\allfiles\undefined
\fi
\ifx\allfiles\undefined

\fi
\section{Experiments}\label{sec:exp}

\subsection{Experimental Settings}\label{sec:3-1}

\textbf{Models.} We utilize the widely adopted Open Pre-trained Transformers (OPTs) \cite{opt} and LLaMAs \cite{llama1,llama2}, with focused model sizes ranging from millions to seven billion parameters. The modules to be pruned are the computationally intensive \texttt{self\_attn.q\_proj}, \texttt{self\_attn.k\_proj}, \texttt{self\_attn.v\_proj}, \texttt{self\_attn.out\_proj}, \texttt{fc1}, \texttt{self\_attn.up\_proj}, \texttt{self\_attn.gate\_proj}, \texttt{fc2} and \texttt{self\_attn.down\_proj} modules constructed from fully-connected layers, which is consistent with baseline.

\textbf{Calibration.} For the data used in calibration, we adhere to the approach outlined in SparseGPT and Wanda, selecting 128 segments of 2048 tokens each from the initial partition of the C4 dataset \cite{c4}. The C4 dataset, sourced from a broad array of internet text, guarantees that our experiments are zero-shot, as no task-specific information is exposed during our optimization process.

\textbf{Datasets and Evaluation.} Regarding evaluation metrics, our primary emphasis is on perplexity, which remains a challenging and reliable metric well suited for evaluating the language modeling capability of compressed models \cite{obc,gptq,zeroquant}. We consider the Raw-WikiText2 \cite{wiki2} test set for perplexity validation. To explore the impact of compression on a broad range of downstream tasks, we also provide zero-shot accuracy results for COPA \cite{copa}, Lambada \cite{lambada}, OpenBookQA \cite{openbookqa}, PIQA \cite{piqa}, RTE \cite{glue}, StoryCloze \cite{storycloze} and Winogrande \cite{winogrande}. 

\textbf{Implementation Details.} We implement the SDS framework in PyTorch \cite{pytorch} and use the HuggingFace Transformers / Datasets library \cite{huggingface} for managing models and datasets. We follow the conventional method (SparseGPT / Wanda) to prune the pre-trained model in the initial pruning step. In the re-dense weight reconstruction step, we use 128 samples as inputs to perform layer-wise knowledge alignment: the number of alignment epochs is 200, the learning rate is 0.1, the loss function is the L2 loss, and the regularization strategy contains L1 and L2 regularization with a ratio of 0.1. The optimization between adjacent layers is achieved by directly using the output of the initial pruned layer as the input for the next layer, eliminating the need for additional forward propagation to obtain the reconstructed layer's output and, thereby, no accumulation of pruning errors. In the second pruning step, we use the same pruning method to prune the re-dense model and use the same configuration as in the previous step without weight regularization to further adjust the weights of the pruned model with a soft sparse mask (exit early and skip weight adjustment when secondary one-shot pruning outperforms initial pruning). The SDS framework uses the same samples throughout while ensuring that no sample is overloaded; it also coincides with the advantage of requiring just a small amount of samples.


\subsection{Results}\label{sec:3-2}


\begin{wraptable}{r}{8cm} \small
\vspace{-1.5em}
\caption{\textbf{Perplexity on Raw-WikiText2 among the SDS workflow.} \textbf{S}{\tiny DS} represents the initially SparseGPT pruned baseline. {\tiny S}\textbf{D}{\tiny S} represents the dense model obtained in the re-dense weight reconstruction step. {\tiny SD}\textbf{S} represents the model obtained in the second round of pruning.}\label{table:ppl}
\begin{center} 
\renewcommand{\arraystretch}{1.2}
\begin{tabular}{cccP{0.6cm}P{0.6cm}P{0.6cm}}
\hlineB{3}
\multirow{2}{*}{PLM} & \multirow{2}{*}{Dense} & \multirow{2}{*}{Sparsity} & \multicolumn{3}{c}{SDS workflow} \\ \cline{4-6} 
 &  &  & \textbf{S}{\tiny DS} & {\tiny S}\textbf{D}{\tiny S} & {\tiny SD}\textbf{S} \\ \hline
\multirow{3}{*}{OPT-125M} & \multirow{3}{*}{27.66} & 50\% & 36.85 & 31.78 &  \textbf{34.23} \\
 &  & 2:4 & 60.43 & 44.46 &  \textbf{51.30} \\
 &  & 4:8 & 44.77 & 37.82 &  \textbf{41.66} \\ \hline
\multirow{3}{*}{OPT-350M} & \multirow{3}{*}{22.01} & 50\% & 31.58 & 24.78 &  \textbf{29.36} \\
 &  & 2:4 & 51.11 & 31.58 &  \textbf{46.23} \\
 &  & 4:8 & 39.59 & 26.15 &  \textbf{34.18} \\ \hline
\multirow{3}{*}{OPT-1.3B} & \multirow{3}{*}{14.62} & 50\% & 17.46 & 17.39 &  \textbf{17.07} \\
 &  & 2:4 & 24.34 & 20.00  &  \textbf{22.67} \\
 &  & 4:8 & 20.05 & 18.06 &  \textbf{19.34} \\ \hline
 \hlineB{3}
\end{tabular}
\end{center}
\end{wraptable}

\textbf{Performance Variations in the SDS Workflow.} \Cref{table:ppl} presents the changes in language modeling perplexity on Raw-WikiText2 after each step of the SDS framework, with SparseGPT as the baseline pruning method and primarily covering several compact OPT models. We focus on three sparsity configurations: 50\% sparsity for model compression, 2:4 and 4:8 sparsity for both compression and real-world computational acceleration on specialized hardware. After the re-dense step ({\tiny S}\textbf{D}{\tiny S}), the perplexity significantly decreases, averaging around 28.0 across all models, which is closer to the dense models (average perplexity of 21.4), with minor performance gap mainly due to regularization effects. Following the second round of pruning ({\tiny SD}\textbf{S}), the performance improves beyond the initial pruning baseline (\textbf{S}{\tiny DS}), averaging around 32.9 compared to the initial average of 36.2. This suggests that the re-dense process successfully produces a more pruning-friendly model.

\textbf{Performance on Language Modeling.} \Cref{table:ppl_all} demonstrates the efficacy of the SDS framework on language modeling with SparseGPT / Wanda as baseline pruning methods. On average, SDS improves perplexity by 1.8 points for the 50\% sparsity level, 7.5 points for the 2:4 sparsity level, and 3.3 points for the 4:8 sparsity level across all models. For example, at 2:4 sparsity, SDS{\tiny -Wanda} reduces perplexity by 39.61 points for OPT-350M, demonstrating the most substantial improvement. These results highlight SDS's ability to produce more pruning-friendly models, significantly enhancing performance compared to the initial pruned baselines.

\begin{table*}[h!]
\caption{\textbf{Language Modeling Performance on Raw-WikiText2.} SparseGPT and Wanda form the base pruning method of the SDS framework, represented as SDS{\tiny-SparseGPT} and SDS{\tiny-Wanda}, respectively. The calibration set utilized is C4.}\label{table:ppl_all}
\begin{center}
\renewcommand{\arraystretch}{1.3}
\setlength{\tabcolsep}{0.9mm}{
\resizebox{\textwidth}{!}{
\begin{tabular}{ccccccccc}
\hlineB{3}
Sparsity & Method & OPT-125M & OPT-350M & OPT-1.3B & OPT-2.7B & LLaMA-7B & LLaMA2-7B \\ \hline
0 & Dense & 27.66 & 22.01 & 14.62 & 12.46 & 5.68 & 5.47 \\
\hline
\multirow{4}{*}{50\%} & SparseGPT & 36.85 & 31.58 & 17.46 & 13.48 & 7.36 & 6.72 \\
                      & \cellr SDS{\tiny -SparseGPT} & \cellr \textbf{34.23} & \cellr \textbf{29.36} & \cellr \textbf{17.07} & \cellr \textbf{13.38} & \cellr \textbf{7.32} & \cellr \textbf{6.69} \\
                      & Wanda & 39.79 & 41.88 & 18.51 & 14.38 & 7.26 & 6.92 \\
                      & \cellr SDS{\tiny -Wanda} & \cellr \textbf{35.05} & \cellr \textbf{33.07} & \cellr \textbf{17.15} & \cellr \textbf{13.70} & \cellr \textbf{7.19} & \cellr \textbf{6.86} \\
\hline
\multirow{4}{*}{2:4} & SparseGPT & 60.43 & 51.11 & 24.34 & 17.18 & 11.32 & 10.88 \\
                     & \cellr SDS{\tiny -SparseGPT} & \cellr \textbf{51.30} & \cellr \textbf{46.23} & \cellr \textbf{22.67} & \cellr \textbf{16.78} & \cellr \textbf{10.65} & \cellr \textbf{10.18} \\
                     & Wanda & 82.47 & 113.17 & 28.33 & 21.20 & 11.54 & 12.14 \\
                      & \cellr SDS{\tiny -Wanda} & \cellr \textbf{59.17} & \cellr \textbf{73.56} & \cellr \textbf{23.94} & \cellr \textbf{17.99} & \cellr \textbf{10.80} & \cellr \textbf{11.34} \\
\hline
\multirow{4}{*}{4:8} & SparseGPT & 44.77 & 39.59 & 20.05 & 14.98 & 8.72 & 8.49 \\
                     & \cellr SDS{\tiny -SparseGPT} & \cellr \textbf{41.66} & \cellr \textbf{34.18} & \cellr \textbf{19.34} & \cellr \textbf{14.81} & \cellr \textbf{8.51} & \cellr \textbf{8.11} \\
                     & Wanda & 53.97 & 62.49 & 22.33 & 16.80 & 8.57 & 8.61 \\
                      & \cellr SDS{\tiny -Wanda} & \cellr \textbf{43.58} & \cellr \textbf{47.31} & \cellr \textbf{19.82} & \cellr \textbf{15.45} & \cellr \textbf{8.39} & \cellr \textbf{8.55} \\
\hlineB{3}
\end{tabular}}}
\end{center}
\end{table*}

\textbf{Performance on Zero-shot Benchmarks.} Downstream tasks offer a more nuanced view, allowing us to test the model's capability across various linguistic and reasoning challenges. \Cref{table:acc} shows the zero-shot performance of the pruned PLMs across multiple downstream tasks. The experimental results demonstrate the superior performance of the SDS framework over baselines. Notably, SDS shows an average improvement of approximately 1.83\% over SparseGPT at a 50\% sparsity level in OPT models. This enhancement is even more significant at the more challenging 2:4 sparsity level, where SDS outperforms SparseGPT by an average of around 2.2\%. This trend of SDS's consistent outperformance is also evident in the LLaMA models. For instance, at 2:4 sparsity, SDS improves accuracy in the LLaMA-7B model to 63.03\% from Wanda's 61.58\% and in the LLaMA2-7B model to 63.51\% from 61.93\%.

\begin{table*}[h]
\caption{\textbf{Multitasking Zero-shot Performance.} Accuracy (\%) was obtained by zero-shot evaluation and averaging over seven downstream tasks, including COPA, Lambada (OPTs only due to the inapplicability of LLaMAs), OpenbookQA, PIQA, RTE, StoryCloze, and Winogrande.}\label{table:acc}
\begin{center}
\renewcommand{\arraystretch}{1.3}
\setlength{\tabcolsep}{0.9mm}{
\resizebox{\textwidth}{!}{
\begin{tabular}{ccccccccc}
\hlineB{3}
Sparsity & Method & OPT-125M & OPT-350M & OPT-1.3B & OPT-2.7B & LLaMA-7B & LLaMA2-7B \\ \hline
0 & Dense & 50.82 & 54.12 & 60.83 & 62.81 & 66.71 & 68.53 \\
\hline
\multirow{4}{*}{50\%} & SparseGPT & 48.85 & 52.33 & 55.89 & 61.14 & 64.75 & 65.78 \\
                      & \cellr SDS{\tiny -SparseGPT} & \cellr \textbf{50.80} & \cellr \textbf{54.51} & \cellr \textbf{58.42} & \cellr \textbf{61.78} & \cellr \textbf{66.16} & \cellr \textbf{66.98} \\
                      & Wanda & 48.46 & 48.90 & 56.18 & 59.36 & 66.26 & 67.08 \\
                      & \cellr SDS{\tiny -Wanda} & \cellr \textbf{49.78} & \cellr \textbf{51.40} & \cellr \textbf{57.58} & \cellr \textbf{60.92} & \cellr \textbf{66.89} & \cellr \textbf{67.58} \\
\hline
\multirow{4}{*}{2:4} & SparseGPT & 47.56 & 48.34 & 53.57 & 58.48 & 62.58 & 64.73 \\
                     & \cellr SDS{\tiny -SparseGPT} & \cellr \textbf{49.61} & \cellr \textbf{50.50} & \cellr \textbf{56.67} & \cellr \textbf{59.96} & \cellr \textbf{63.22} & \cellr \textbf{65.43} \\
                     & Wanda & 45.69 & 44.77 & 52.86 & 55.51 & 61.58 & 61.93 \\
                      & \cellr SDS{\tiny -Wanda} & \cellr \textbf{47.09} & \cellr \textbf{46.69} & \cellr \textbf{54.44} & \cellr \textbf{58.98} & \cellr \textbf{63.03} & \cellr \textbf{63.51} \\
\hline
\multirow{4}{*}{4:8} & SparseGPT & 48.29 & 49.85 & 54.95 & 60.24 & 64.40 & 65.19 \\
                     & \cellr SDS{\tiny -SparseGPT} & \cellr \textbf{49.67} & \cellr \textbf{52.25} & \cellr \textbf{57.92} & \cellr \textbf{61.48} & \cellr \textbf{65.61} & \cellr \textbf{66.05} \\
                     & Wanda & 46.28 & 46.41 & 55.04 & 58.21 & 64.60 & 66.21 \\
                      & \cellr SDS{\tiny -Wanda} & \cellr \textbf{47.70} & \cellr \textbf{48.61} & \cellr \textbf{56.12} & \cellr \textbf{59.91} & \cellr \textbf{65.65} & \cellr \textbf{66.42} \\
\hlineB{3}
\end{tabular}}}
\end{center}
\end{table*}

\begin{table*}[h]
\begin{center}
\caption{\textbf{Comparison of Different Configurations of the SDS Framework.} We compare the language understanding perplexity and accuracy of OPT-125m on eight tasks in a 2:4 sparse configuration with SparseGPT as the base pruning method. The {\color{lightgray}gray characters} represent the skipped steps; DD stands for dense data, which uses the activations generated by the dense model as inputs for weight adjustment; SD stands for sparse data, which uses the activations generated by the sparse model as inputs for weight adjustment; KD stands for KD-aware data, which uses the activations of the model after weight adjustment as inputs for the next layer of weight adjustment; WR represents weight regularization; MSD stands for multiple sparse data, which means that different samples are used for each step of the SDS process.}
\label{table:abl}
\renewcommand{\arraystretch}{1.3} 
\setlength{\tabcolsep}{1mm}{
\resizebox{\textwidth}{!}{
\begin{tabular}{rcccccccccc}
\hlineB{3}
& Method & Wiki.↓ & COPA↑ & Lamb.↑ & BookQ.↑ & PIQA↑ & RTE↑ & Story.↑ & Wino.↑ & Avg acc.↑ \\ \hline
{\small 1:} & \textbf{Dense} & 27.66 & 66 & 39.16 & 28.0 & 62.02 & 50.18 & 60.03 & 50.36 & 50.82\\
{\small 2:} & \textbf{S}{\color{lightgray}\tiny{DS}} & 60.43 & 62 & 27.55 & 25.8 & 57.24 & 53.79 & 55.38 & 51.14 & 47.56 \\ \hline
{\small 3:} & {\color{lightgray}\tiny{SD}}\textbf{S} {\small w DD} & 58.63 & 62 & 27.03 & 26.0 & 58.11 & 52.35 & 55.25 & 50.75 & 47.36\\
{\small 4:} & {\color{lightgray}\tiny{SD}}\textbf{S} {\small w SD} & 58.56 & 62 & 20.96 & 26.4 & 58.65 & 51.62 & 56.21 & 50.98 & 46.69\\
{\small 5:} & {\color{lightgray}\tiny{SD}}\textbf{S} {\small w KD} & 56.82 & 63 & 30.04 & 26.2 & 58.76 & 53.79 & 55.63 & 49.64 & 48.15 \\ \hline
{\small 6:} & {\color{lightgray}\tiny{S}}\textbf{DS} & 57.98 & 62 & 26.68 & 26.2 & 59.30 & 48.74 & 56.27 & 50.98 & 47.17 \\ \hline
{\small 7:} & \textbf{SDS} {\small w/o WR} & 51.96 & 63 & 30.04 & 26.2 & 58.92 & 51.95 & 56.46 & 51.38 & 48.29\\
{\small 8:} & \textbf{SDS} {\small w DD} & 57.72 & 62 & 26.99 & 26.0 & 59.30 & \textbf{54.15} & 55.19 & \textbf{51.46} & 47.87\\
{\small 9:} & \textbf{SDS} {\small w KD} & 57.32 & 61 & 29.15 & 26.4 & 59.51 & 54.01 & 54.17 & 51.38 & 47.95\\
\gr {\small 10:}  & \textbf{SDS} {\small w SD} & \textbf{51.30} & \textbf{65} & \textbf{31.57} & \textbf{27.8} & 59.85 & \textbf{54.15} & \textbf{57.42} & 51.46 & \textbf{49.61} \\ \hline
{\small 11:} & \textbf{SDS} {\small w MSD} & 52.06 & 64 & 30.35 & 27.0 & \textbf{60.01} & 50.18 & 57.16 & \textbf{52.57} & 48.75 \\ \hlineB{3}
\end{tabular}}}
\end{center}
\end{table*}

In summary, our evaluations convincingly demonstrate the robustness and efficacy of the SDS framework across a variety of sparsity configurations. Both language modeling and zero-shot downstream multitask performance metrics affirm the consistent superiority of SDS over baselines. Therefore, SDS is an effective pruning method for PLMs.

\enlargethispage{\baselineskip}

\subsection{Ablation Study}\label{sec:3-3}

To validate the effectiveness of \textit{the step composition} and \textit{the sparse regularization} of the Sparse-Dense-Sparse (SDS) framework, we conducted a series of ablation experiments as shown in \Cref{table:abl}. The first two rows represent the dense and the SparseGPT pruned baselines, respectively. 

Rows 3 to 5 verify the effect of only performing one-shot pruning and sparse weight adjustment. This approach reflects the effect of pruning when the loss is formally computed instead of the "Hessian approximation". Overall, {\color{lightgray}\small{SD}}\textbf{S} was only able to outperform SparseGPT on three tasks completely. Comparing the different input data used in the {\color{lightgray}\small{SD}}\textbf{S} case, {\color{lightgray}\small{SD}}\textbf{S} w KD can outperform SparseGPT on seven tasks, which is considered better than choosing the other two data types. Thus it can be concluded that \textit{{\color{lightgray}\small{SD}}\textbf{S} mode has limited optimization for SparseGPT} and \textit{selection of data with low loss (KD) is more suitable for {\color{lightgray}\small{SD}}\textbf{S} mode than selection of data with high loss (DD or SD)}.

Row 6 verifies the effect of the second round pruning of the dense model after injecting it directly with sparse regularization, skipping the initial pruning, i.e., residual sparse characteristics. {\color{lightgray}\small{S}}\textbf{DS} outperforms the performance of SparseGPT on five tasks, but it does not yet reach the superior performance of \textbf{SDS} w SD. This observation demonstrates that \textit{residual sparse characteristics are effective}.

Rows 7 to 10 verify the role of weight-based and data-based regularization in \textbf{SDS}, respectively. Unlike {\color{lightgray}\small{SD}}\textbf{S}, SD is a more suitable data choice for \textbf{SDS}, and this harder data serves the purpose of regularization while avoiding the challenge of learning hard data in multiple steps. Also, it can be argued that residual sparse characteristics and data regularization dominate in sparse regularization compared to weight regularization. \Cref{app:3} provides an analysis from a distributional perspective.

Row 11 shows the impact of using different samples at each step of the SDS process. The optimization is closer to SDS w SD, but only two tasks outperform it. This indicates that SDS achieves favorable results and does not necessitate additional samples.

\subsection{Efficiency Analysis}
To illustrate the enhanced efficiency of pruned models, we present the inference speed of dense and sparse models on AMD CPU. We use the DeepSparse library \cite{deepsparse} and apply 50\% unstructured pruning on OPTs and LLaMAs in this experiment. Table \ref{tab:amdcpu} indicates that the pruned models can achieve 1.19x $\sim$ 1.87x speedup compared to their dense counterparts. This significant boost in inference speed underscores the critical importance of model pruning in practical applications. 

\begin{table}[h]
\centering
\renewcommand{\arraystretch}{1.3}
\caption{\textbf{Inference Speed Comparison of Pre- and Post-pruned PLMs} using DeepSparse on AMD Ryzen 7 PRO 5850U @ 1.90 GHz with batchsize = 1 and sequence\_length = 2048, throughput (batch/sec) and latency (ms/batch) are tested.}\label{tab:amdcpu}
\setlength{\tabcolsep}{0.4mm}{
\vspace{3mm}\small
\resizebox{\textwidth}{!}{
\begin{tabular}{c|cc|cc|cc|cc}
\hlineB{3}
Model & \multicolumn{2}{c|}{OPT-1.3B} & \multicolumn{2}{c|}{OPT-2.7B} & \multicolumn{2}{c|}{OPT-6.7B} & \multicolumn{2}{c}{LLaMA-7B} \\ \hline
Metric & Throughput & Latency & Throughput & Latency & Throughput & Latency & Throughput & Latency \\ \hline
Dense & 14.1 & 71.19{\small ±1.5} & 6.9 & 145.17{\small ±3.1} & 2.9 & 345.39{\small ±7.1} & 2.3 & 409.98{\small ±5.5} \\ \hline
Sparse & 16.8 & 59.55{\small ±2.1} & 9.3 & 107.07{\small ±1.9} & 4.4 & 225.67{\small ±2.8} & 4.3 & 229.52{\small ±2.3} \\ \hline
Speedup & \multicolumn{2}{c|}{\textbf{1.19x}} & \multicolumn{2}{c|}{\textbf{1.35x}} & \multicolumn{2}{c|}{\textbf{1.52x}} & \multicolumn{2}{c}{\textbf{1.87x}} \\ \hlineB{3}
\end{tabular}}}
\end{table}

\ifx\allfiles\undefined
\fi
\ifx\allfiles\undefined

\fi
\section{Related Work}\label{sec:related_work}

\textbf{Pruning for Language Model Compression}. The surging complexity of Transformer-based language models, which now feature hundreds of billions of parameters, has accentuated the urgent need for effective and efficient model pruning methods \cite{deepcompression,han2015learning, obs}. 
These pruning methods can be broadly classified into structured and unstructured approaches.
Structured pruning is more hardware-friendly, as it directly prunes entire segments of weights, thereby removing consecutive computations \cite{llm_pruner, dejavu}.
Unstructured pruning (sparsification) is also receiving interest, particularly as hardware advancements increasingly support the acceleration of sparse patterns such as 2:4 or 4:8 sparse \cite{nmsparse}.
Techniques such as SparseGPT \cite{sparsegpt} extend the OBS \cite{obs} methodology to column-wise prune weights, allowing the modification of values in the unpruned columns to compensate for the pruning errors.
Aaquib et al. \cite{prune_and_tune} enhance SparseGPT by incorporating minimal iterative task fine-tuning during the pruning process, demonstrating performance improvements at high sparsity levels.
Wanda \cite{wanda} introduces a simple yet effective no-retraining-needed pruning strategy that prunes weights based on their magnitudes and corresponding activations. 
OWL \cite{owl} considers the inhomogeneous distribution of interlayer outliers and further extends Wanda to a non-uniform sparsity distribution, achieving better performance.
DS$\varnothing$T \cite{dsot} and SPP \cite{spp}, as fine-tuning methods designed for sparse models, can improve the performance of pruned PLMs within limited complexity.

\textbf{Weight Distribution Optimization}. Various techniques have been employed to understand and optimize weight distributions in the quest for more efficient neural networks. 
The Dense-Sparse-Dense training method \cite{dsd} provides a three-step flow: an initial dense training to learn connection weights, a sparsity-inducing phase that prunes unimportant connections, and a final re-dense step. This process improves performance across various network architectures and underscores the importance of parameter distribution in achieving better local optima. Ji et al. \cite{dsd_pro} treat the process of Dense-Sparse-Dense as a regularized entirety, improving the model's cross-domain few-shot classification capability. Regularization methods serve as pivotal tools for optimizing the parameter distribution. SPDF \cite{SPDF} adopts a language model construction paradigm of sparse pre-training and dense fine-tuning, which both introduce sparse regularization to the training process and improve the training efficiency. Dropout \cite{dropout} is a form of ensemble learning of neural networks. It implicitly changes the parameter distribution by randomly zeroing out weights during training, encouraging a sparse representation. Yoshida et al. \cite{sp_reg} focus on constraining the spectral norm of weights matrices to improve the generalization capabilities of neural networks. This method plays a crucial role in shaping the parameter space, making it more amenable to sparse approximations.

In this paper, the proposed Sparse-Dense-Sparse (SDS) framework first regularizes the weights into a pruning-friendly dense distribution and prunes the models, aiming to enhance the language comprehension and multitasking performance of the state-of-the-art conventional pruning methods.

\ifx\allfiles\undefined
\fi
\ifx\allfiles\undefined

\fi

\section{Limitation}\label{limitation}

The Sparse-Dense-Sparse (SDS) framework improves pruned PLMs through the perspective of weight distribution optimization. Our approach surpasses the previous SOTA methods, including SparseGPT and Wanda, under the same sparsity configuration. One limitation is that our sparsity optimization process consumes more computation overhead (it takes more than two hours to optimize a seven-billion scale model in parallel on eight GPUs.). However, the introduced optimization time is still tiny compared to training a language model. Once the sparse model is obtained and deployed, one can benefit from the acceleration on the specific hardware.

\section{Conclusion}\label{sec:conclusion}

We introduced the Sparse-Dense-Sparse (SDS) framework for optimizing pruned pre-trained language models (PLMs), consisting of initial pruning, re-dense weight reconstruction, and a second pruning round. The SDS framework focuses on weight distribution optimization and incorporates sparse regularization elements—including residual sparse characteristics, data-based regularization, and weight-based regularization. As a result, SDS not only enhances the model's pruning friendliness but also achieves state-of-the-art pruning results.
Experimental results show that SDS reduces perplexity by 9.13 on Raw-Wikitext2 and enhances accuracy by an average of 2.05\% across various zero-shot benchmarks for OPT-125M with a 2:4 sparsity configuration.

\ifx\allfiles\undefined
\fi

\newpage
\appendix
\onecolumn
\ifx\allfiles\undefined

\fi
\section{Appendix}\label{sec:conclusion}

\subsection{Error Accumulation and Data-based Regularization}\label{app:2}

\begin{wrapfigure}{r}{0.64\textwidth}
  \centering
  \begin{minipage}{0.31\textwidth}
    \includegraphics[width=\textwidth]{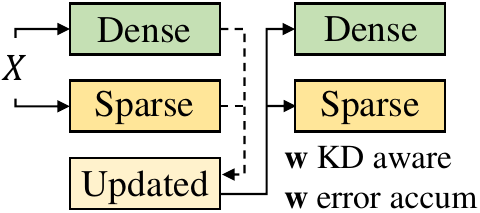} \\
    \centering{(a) KD-data}
  \end{minipage}
  \hfill
  \begin{minipage}{0.31\textwidth}
    \includegraphics[width=\textwidth]{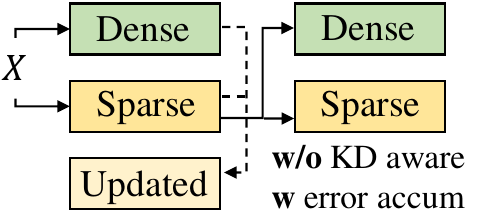}\\
    \centering{(b) SD-data}
  \end{minipage}
  \\
  \begin{minipage}{0.31\textwidth}
    \includegraphics[width=\textwidth]{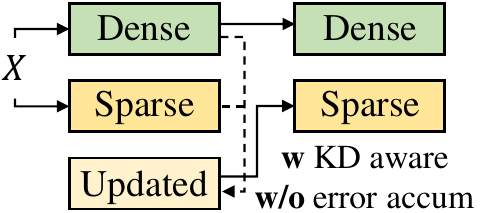}\\
    \centering{(c) DD-data}
  \end{minipage}
  \hfill
  \begin{minipage}{0.31\textwidth}
    \includegraphics[width=\textwidth]{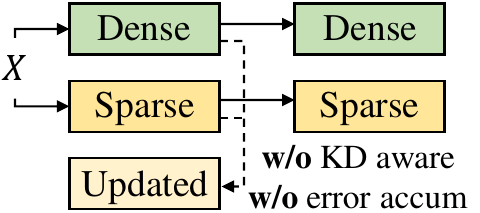}\\
    \centering{(d) DD-data}
  \end{minipage}
  \caption{\textbf{Four Data Selection Paradigms in Weight Adjustment.} Straight lines represent forward propagation, and dashed lines represent knowledge distillation.}\label{fig:data}
  \vspace{-1em}
\end{wrapfigure}

The input data used in weight adjustment can be categorized in two ways: whether to perform error accumulation and whether to be aware of the knowledge distillation (KD) process. \Cref{fig:data} presents four different data selection ways for weight adjustment. 

\textbf{(a)} Weight adjustment with KD aware and error accumulation, this paradigm corresponds to \textit{KD-data} in our ablation study (cf., \Cref{sec:3-3}). After applying KD to the sparse layer, a subsequent forward propagation is needed to generate inputs for the next layer. These inputs are solely based on the former layer's outputs, thus accumulating errors. Since KD aims to reduce loss, this extra forward propagation simplifies the data, making it easier for the subsequent layer to learn.
\textbf{(b)} Weight adjustment with error accumulation but without KD aware, this paradigm corresponds to \textit{SD-data} in our ablation study. Unlike paradigm (a), this approach abandons the additional forward propagation to account for changes in the layer updated by KD. This results in the next layer of learning from data corresponding to a higher loss, making learning more challenging than in paradigm (a).
\textbf{(c)} and \textbf{(d)} are two ways of adjusting the weights without accumulating errors. The presence or absence of KD awareness has a minimal impact on either, as the optimization direction is constrained by the same dense model in both cases. The \textit{DD-data} paradigm in our ablation study employs paradigm (d).

From the perspective of data difficulty, \textit{DD-data} is the most difficult because it requires each layer to compensate for the errors accumulated in all previous layers. This difficulty is more prominent in the KD process under sparsity constraints. In the ablation study (cf., \Cref{sec:3-3}), neither one nor two times optimization of the sparse model using \textit{DD-data} was able to achieve excellent results, verifying the above observation. \textit{KD-data} is the easiest because the weights of the sparse model are updated in the direction of lower loss during knowledge distillation. The use of \textit{KD-data} has yielded relatively good results only in single-step optimization of the sparse model since simple data carries less data regularization and a relatively low upper bound for optimization. \textit{SD-data} is relatively moderate in difficulty and comes with data regularization and hence achieved an ideal result in SDS's optimization of the sparse model. The reason why \textit{SD-data} did not achieve an ideal result in the single-step optimization could be the challenge of the difficult data. 




\subsection{Distribution Analysis}\label{app:3}

\begin{figure}[h!]
\centering
    \includegraphics[width=0.95\textwidth]{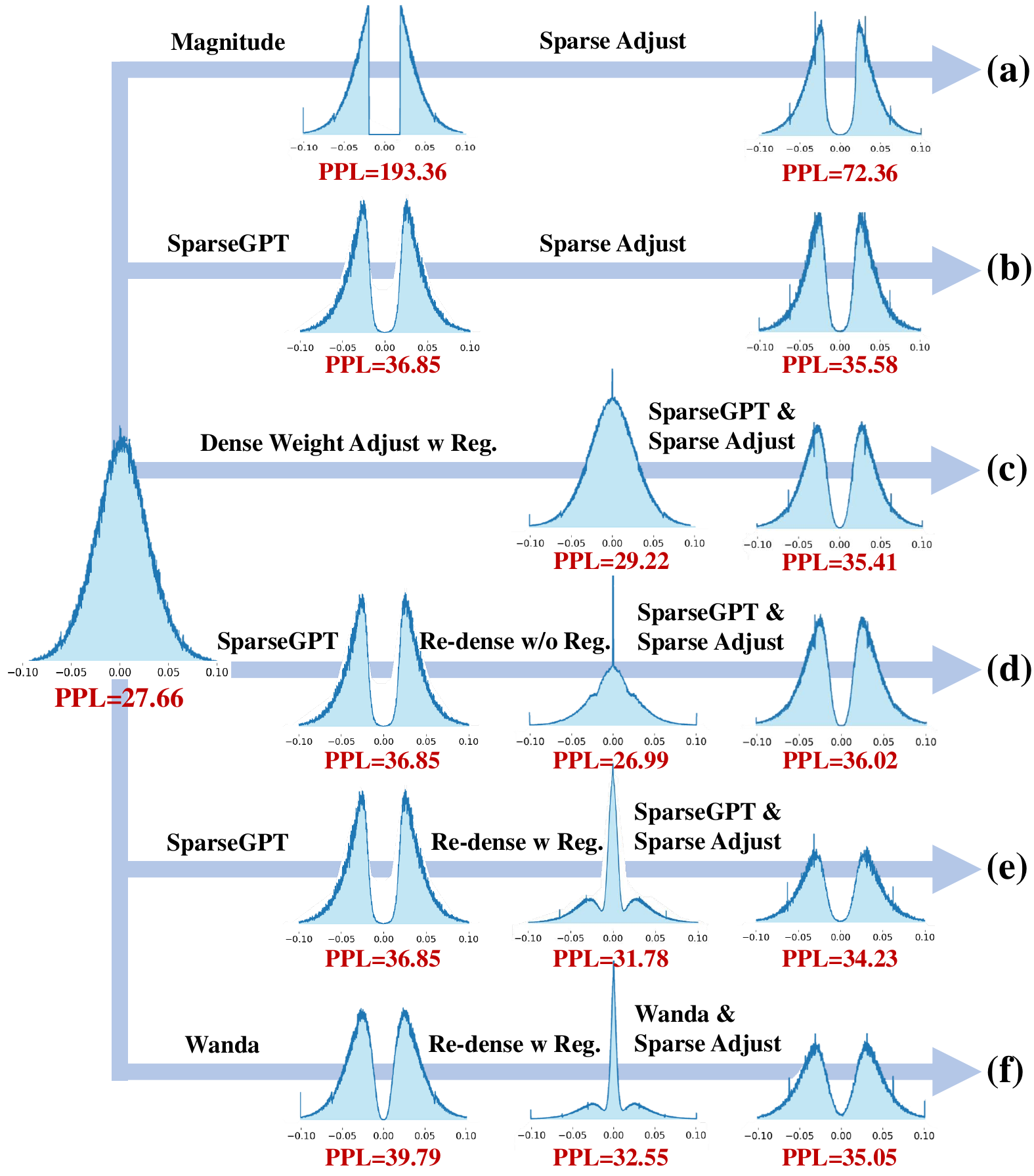} 
    \caption{\textbf{Changes in Distributions During Optimization of Pruned PLMs.} The distribution observations are from the last layer of OPT-125m with 50\% pruning. \textbf{(a)} represents the process of first pruning the model by magnitude (\textit{absmin}) \cite{magnitude} and then optimizing the pruned model using SD-data. \textbf{(b)} represents the {\color{lightgray}\small{SD}}\textbf{S} w KD in the ablation study (cf., \Cref{sec:3-3}). \textbf{(c)} represents the {\color{lightgray}\small{S}}\textbf{DS}. \textbf{(d)} represents the \textbf{SDS} w KD. \textbf{(e)} represents the \textbf{SDS} w SD. \textbf{(f)} represents the \textbf{SDS} w SD and with Wanda as the pruning method. Zero values are omitted in sparse weight distributions for better clarity.}
    \label{fig:dist_all}
\end{figure}

\Cref{fig:dist_all} visualizes the impact of several pertinent optimizations performed on pruned PLMs from the perspective of distribution changes.

Magnitude-based one-shot pruning method is ineffective on PLMs primarily because it focuses only on the absolute value of the weights. This simplistic approach tends to create a truncated bimodal distribution of the model weights, concentrating them at extreme positive and negative values. Distribution truncation can lead to model instability, as removing near-zero weights disrupts the model's ability to make subtle, nuanced adjustments. Due to the large amount of information lost in the pruning process, the model's performance can only be recovered to a limited extent after weight adjustment. In contrast, modern pruning methods like SparseGPT consider higher-order rather than zero-order information, which manages to maintain an untruncated bimodal distribution similar to what magnitude pruning plus subsequent weight adjustment would achieve. However, they do it in a single step and can achieve better performance. 

As shown in \Cref{fig:dist_all}b, the model has a relatively sharp bimodal peak in its distribution after being pruned by SparseGPT, which challenges the model's generalization ability, optimization space, and stability. Direct adjustment of the pruned model's weights yields limited performance and optimization of the weight distribution. Therefore, it is necessary to consider the SDS process. 

Before attempting the SDS process, \Cref{fig:dist_all}c and \Cref{fig:dist_all}d show the trend of weight distribution changes for only injecting weight regularization or residual sparse characteristics into the model, respectively. Both find a new dense solution to some extent: a dense model with a smoother distribution and more zeros can be found by using data-based regularization and weight-based regularization for dense weight adjustment, and a dense model that converges to a multi-peaked distribution with more zeros is obtained after re-dense reconstruction of the sparse model without regular regularization. After a second round of pruning, both approaches lead to a recovery in the model's performance. However, they are less effective than the SDS process that uses a combination of data-based regularization, weight-based regularization, and residual sparse characteristics, as shown in \Cref{fig:dist_all}e and \Cref{fig:dist_all}f. This illustrates the effectiveness and mutual reinforcement effect of regularization techniques in the SDS framework. An interesting phenomenon is that when regular regularization is not used, it is possible to reconstruct the pruned model to equal or even higher performance than the original dense model. This is perhaps due to the absence of regularization techniques, which allowed the re-dense model to overfit the behavior of the original dense model. The limited performance improvement of the re-dense model after a second round of pruning also supports the above deduction.

\subsection{SDS for Uniform and Non-uniform Sparsity}\label{uniform}
Uniform sparsity implies that each layer has the same sparsity configuration, while non-uniform sparsity allows for different levels of sparsity in different layers. In Section \ref{sec:exp}, we apply uniform sparsity to PLMs using 50\%, 2:4, or 4:8 configurations. In this section, we test various sparse configurations to verify the robustness of SDS.

\begin{figure*}[hb]
    \centering
    \begin{minipage}{0.45\textwidth}
    \centering
    \hspace{0.9cm}OPT-125M \\
    \includegraphics[width=\textwidth]{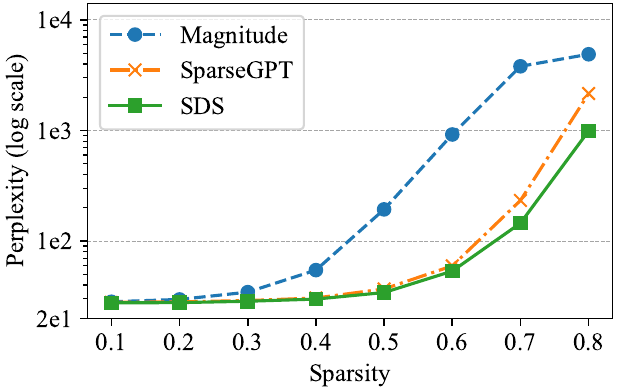} 
  \end{minipage}
  \hfill
  \begin{minipage}{0.45\textwidth}
   \centering
    \hspace{0.9cm}OPT-350M \\
    \includegraphics[width=\textwidth]{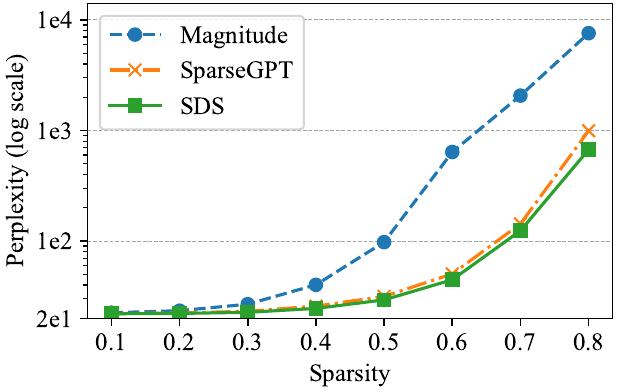}
  \end{minipage}
    \caption{\textbf{Sparsity vs. Perplexity} in OPTs.}
    \label{fig:sps_vs_ppl}
\end{figure*}

\textbf{Varying Sparsity Levels.} We conduct experiments with varying uniform sparsity levels for unstructured pruning; the results are depicted in \Cref{fig:sps_vs_ppl}. It can be observed that the Sparse-Dense-Sparse (SDS) framework is effective in optimizing the performance of the pruned PLMs at either high or low sparsity.

\textbf{Non-uniform Sparsity.} Outlier Weighted layer-wise Sparsity (OWL) \cite{owl}, which guides layer sparsity distribution optimization by examining the frequency of outliers in PLMs. By incorporating OWL as the basic pruning method within our SDS framework, we have successfully optimized performance for both the OPT-125M and OPT-1.3B models. \Cref{tab:uniform} illustrates the significant improvements achieved through our SDS-OWL integration, compared to the baseline OWL-SparseGPT and OWL-Wanda methods.

\begin{table}[h!]
\centering
\renewcommand{\arraystretch}{1.3}
\caption{\textbf{SDS Optimization Effect on OWL Non-uniform Sparsity.} Perplexity is evaluated on Raw-Wikitext2. Accuracy (\%) is obtained by zero-shot evaluation and averaging over seven downstream tasks, including COPA, Lambada, OpenbookQA, PIQA, RTE, StoryCloze, and Winogrande.}\label{tab:uniform}
\resizebox{0.9\textwidth}{!}{
\begin{tabular}{lccccc}
\hlineB{3}
\multirow{2}{*}{Method} & \multirow{2}{*}{Sparsity} & \multicolumn{2}{c}{OPT-125M}                    & \multicolumn{2}{c}{OPT-1.3B}                    \\ \cline{3-6} 
                        &                           & PPL$\downarrow$           & AVG Acc\%$\uparrow$         & PPL$\downarrow$           & AVG Acc\%$\uparrow$         \\ \hline
OWL-SparseGPT           & 0.5                       & 36.71         & 48.56             & 17.58         & 57.2              \\
\gr SDS-OWL-SparseGPT       & 0.5                       & \textbf{34.97}         & \textbf{50.39}             & \textbf{17.39}         & \textbf{57.97}             \\ \hline
OWL-SparseGPT           & 0.7                       & 199.34        & 43.67             & 50.47         & 50.35             \\
\gr SDS-OWL-SparseGPT       & 0.7                       & \textbf{161.75}        & \textbf{44.06}             & \textbf{42.14}        & \textbf{52.27}             \\ \hline
OWL-Wanda               & 0.5                       & 39.09         & 48.81             & 18.48         & 56.35             \\
\gr SDS-OWL-Wanda           & 0.5                       & \textbf{35.26}         & \textbf{49.36}             & \textbf{17.52}         & \textbf{58.33}             \\ \hline
OWL-Wanda               & 0.7                       & 303.18        & 41.81             & 102.88        & 45.44             \\
\gr SDS-OWL-Wanda           & 0.7                       & \textbf{201.1}         & \textbf{43.06}             & \textbf{70.96}         & \textbf{47.65}             \\ \hlineB{3}
\end{tabular}
}
\end{table}

\ifx\allfiles\undefined
\fi


\end{document}


\fi
\section{Appendix}\label{sec:conclusion}

\subsection{Additional experiments}\label{app:5}

\textbf{General applicability of SDS.} To verify the general applicability of the Sparse-Dense-Sparse framework to other pruning methods and models, we additionally chose to perform experimental validation on the GPT2 \cite{gpt2} models and the wanda pruning method.

Among them, the GPT2 model covers four versions of model instances, including Small (S), Medium (M), Large (L), and XLarge (XL) ones, with parameter sizes ranging from 124M to 1.5B approximately.

The Wanda pruning method considers both weights and activations as a saliency measure $\mathbf{S}$ for finding the position of the sparse mask:

\begin{equation}
    \mathbf{S}_{i j}=\left|\mathbf{W}_{i j}\right| \cdot\left\|\mathbf{X}_j\right\|_2,
\end{equation}

where $|\cdot|$ denotes the absolute value operator, and $\left\|\mathbf{X}_j\right\|_2$ computes the $\ell2$ norm of the $j$th features gathered across token dimension. The final saliency score is ascertained by the multiplication of these two scalar values.
Compared to SparseGPT, Wanda is able to achieve large model pruning similar to SparseGPT without weight modification, which contributes to the simplicity of Wanda.

Tables \ref{tab:total_ppl} and \ref{tab:total_acc_avg} show the results of the complete language modeling perplexity and downstream multi-task zero-shot experiments, respectively.

Based on the results of the empirical evaluation, it is evident that the Sparse-Dense-Sparse (SDS) approach significantly outperforms both SparseGPT and Wanda methods across multiple sparsity configurations and model sizes. A brief critique of the alternative methods reveals that SparseGPT often shows higher perplexity scores and lower multitask zero-shot accuracy across configurations. Wanda, on the other hand, produces even poorer results on these undersized PLMs, as evidenced by the prohibitively high perplexity observed in the OPT-350m model with 2:4 sparse configuration.  

In terms of language modeling perplexity, SDS demonstrates a clear advantage, improving on average by around 11\% over SparseGPT across various sparsity configurations and model sizes. It also shows significant improvements over Wanda, with an average gain of roughly 18\% in perplexity metrics. In the multitask zero-shot experiment, the superiority of SDS is further evidenced; it outperforms SparseGPT by an average of 2.5\% and surpasses Wanda by an average of 3.1\%. For details of the zero-shot experiments on individual tasks, refer to Tables \ref{tab:copa} to \ref{tab:winogrande}.

Overall, these findings confirm the superiority of the SDS framework, not only in terms of preserving model generalization capabilities but also in improving them, making it an efficient and effective strategy for model pruning and performance optimization.

\begin{table}[t]
\centering
\caption{\textbf{Perplexity on Wikitext2.} SparseGPT and wanda were used to compose the SDS framework, denoted as SDS{\tiny-SparseGPT} and SDS{\tiny-Wanda}, respectively. the calibration set used was C4.}
\label{tab:total_ppl}
\setlength{\tabcolsep}{0.9mm}{
\renewcommand{\arraystretch}{1.2} 
\begin{tabular}{cccccccccc}
\hline
Sparsity & Method & OPT-125m & opt-350m & opt-1.3b & opt-2.7b & gpt2-S & gpt2-M & gpt2-L & gpt2-XL \\ \hline
0 & - & 27.66 & 22.01 & 14.62 & 12.16 & 29.95 & 21.71 & 19.33 & 17.41 \\ \hline
\multirow{4}{*}{50\%} & SparseGPT & 36.85 & 31.58 & 17.46 & 13.48 & 47.46 & 28.06 & 23.55 & 19.91 \\
 & SDS\tiny{-SparseGPT} & \textbf{34.23} & \textbf{29.36} & \textbf{17.40} & \textbf{13.42} & \textbf{45.76} & \textbf{27.91} & \textbf{23.32} & \textbf{19.62} \\ \cline{2-10} 
 & Wanda & 39.79 & 41.88 & 18.36 & 14.38 & 46.71 & 29.29 & 24.89 & 20.83 \\
 & SDS\tiny{-Wanda} & \textbf{35.05} & \textbf{33.07} & \textbf{17.23} & \textbf{13.74} & \textbf{40.32} & \textbf{27.39} & \textbf{23.15} & \textbf{19.78} \\ \hline
\multirow{4}{*}{2:4} & SparseGPT & 60.43 & 51.11 & 23.90 & 17.18 & 73.11 & 40.41 & 32.49 & 25.97 \\
 & SDS\tiny{-SparseGPT} & \textbf{51.30} & \textbf{46.23} & \textbf{23.02} & \textbf{17.36} & \textbf{64.31} & \textbf{38.24} & \textbf{31.33} & \textbf{25.05} \\ \cline{2-10} 
 & Wanda & 82.47 & 113.17 & 27.32 & 20.94 & 123.66 & 61.70 & 52.39 & 32.60 \\
 & SDS\tiny{-Wanda} & \textbf{59.17} & \textbf{73.56} & \textbf{23.94} & \textbf{18.02} & \textbf{63.57} & \textbf{41.11} & \textbf{31.11} & \textbf{25.35} \\ \hline
\multirow{4}{*}{4:8} & SparseGPT & 44.77 & 39.59 & 19.95 & 14.98 & 53.14 & 32.84 & 26.77 & 22.70 \\
 & SDS\tiny{-SparseGPT} & \textbf{41.66} & \textbf{34.18} & \textbf{19.54} & \textbf{14.81} & \textbf{50.90} & \textbf{32.41} & \textbf{26.29} & \textbf{22.27} \\ \cline{2-10} 
 & Wanda & 53.97 & 62.49 & 21.96 & 16.80 & 73.73 & 41.12 & 32.58 & 25.14 \\
 & SDS\tiny{-Wanda} & \textbf{43.58} & \textbf{47.31} & \textbf{19.82} & \textbf{15.45} & \textbf{51.05} & \textbf{33.55} & \textbf{26.16} & \textbf{22.11} \\ \hline
\end{tabular}}
\end{table}

\begin{table}[t]
\centering
\caption{\textbf{Multitasking zero-shot accuracy comparison.} Accuracy (\%) was obtained by zero-shot evaluation and averaging over seven downstream tasks, including COPA, Lambada, OpenbookQA, PIQA, RTE, StoryCloze, and Winogrande.}
\label{tab:total_acc_avg}
\setlength{\tabcolsep}{0.9mm}{
\renewcommand{\arraystretch}{1.2} 
\begin{tabular}{cccccccccc}
\hline
Sparsity & Methpd & opt-125m & opt-350m & opt-1.3b & opt-2.7b & gpt2-S & gpt2-M & gpt2-L & gpt2-XL \\ \hline
0 & - & 50.82 & 54.12 & 60.83 & 62.81 & 50.07 & 53.58 & 56.51 & 58.40 \\ \hline
\multirow{4}{*}{50\%} & SparseGPT & 48.85 & 52.33 & 55.89 & 61.14 & 47.27 & 52.82 & 53.47 & 57.22 \\
 & SDS\tiny{-SparseGPT} & \textbf{50.80} & \textbf{54.51} & \textbf{58.42} & \textbf{61.78} & \textbf{48.37} & \textbf{53.33} & \textbf{54.34} & \textbf{58.00} \\ \cline{2-10} 
 & Wanda & 48.46 & 48.90 & 56.18 & 59.36 & 46.50 & 52.01 & 53.65 & 56.03 \\
 & SDS\tiny{-Wanda} & \textbf{49.78} & \textbf{51.40} & \textbf{57.58} & \textbf{60.92} & \textbf{49.05} & \textbf{53.34} & \textbf{54.88} & \textbf{57.36} \\ \hline
\multirow{4}{*}{2:4} & SparseGPT & 47.56 & 48.34 & 53.57 & 58.48 & 46.47 & 50.17 & 50.85 & 53.57 \\
 & SDS\tiny{-SparseGPT} & \textbf{49.59} & \textbf{50.50} & \textbf{56.67} & \textbf{59.96} & \textbf{47.62} & \textbf{50.65} & \textbf{52.45} & \textbf{56.25} \\ \cline{2-10} 
 & Wanda & 45.69 & 44.77 & 52.86 & 55.51 & 42.32 & 47.38 & 48.92 & 51.37 \\
 & SDS\tiny{-Wanda} & \textbf{47.09} & \textbf{46.69} & \textbf{54.44} & \textbf{58.98} & \textbf{46.55} & \textbf{51.48} & \textbf{52.77} & \textbf{54.68} \\ \hline
\multirow{4}{*}{4:8} & SparseGPT & 48.29 & 49.85 & 54.94 & 60.24 & 46.32 & 51.04 & 52.53 & 55.77 \\
 & SDS\tiny{-SparseGPT} & \textbf{49.67} & \textbf{52.25} & \textbf{57.92} & \textbf{61.48} & \textbf{48.00} & \textbf{52.15} & \textbf{53.45} & \textbf{55.85} \\ \cline{2-10} 
 & Wanda & 46.28 & 46.41 & 55.04 & 58.21 & 44.63 & 49.15 & 51.17 & 54.18 \\
 & SDS\tiny{-Wanda} & \textbf{47.70} & \textbf{48.61} & \textbf{56.12} & \textbf{59.91} & \textbf{47.07} & \textbf{52.04} & \textbf{54.44} & \textbf{56.90} \\ \hline
\end{tabular}}
\end{table}

\begin{table}[t]
\centering
\caption{Zero-shot performance (accuracy \%) on COPA}
\label{tab:copa}
\setlength{\tabcolsep}{0.9mm}{
\renewcommand{\arraystretch}{1.2} 
\begin{tabular}{cccccccccc}
\hline
Sparsity & Method & opt-125m & opt-350m & opt-1.3b & opt-2.7b & gpt2-S & gpt2-M & gpt2-L & gpt2-XL \\ \hline
0 & - & 66 & 72 & 79 & 77 & 64 & 64 & 72 & 73 \\ \hline
\multirow{4}{*}{50\%} & SparseGPT & 64 & 68 & 71 & \textbf{77} & 60 & 64 & 65 & 74 \\
 & SDS\tiny{-SparseGPT} & \textbf{68} & \textbf{69} & \textbf{78} & 75 & \textbf{61} & \textbf{66} & \textbf{67} & \textbf{75} \\ \cline{2-10} 
 & Wanda & 62 & 65 & \textbf{73} & 70 & 60 & 68 & 68 & 72 \\
 & SDS\tiny{-Wanda} & \textbf{64} & \textbf{66} & 71 & \textbf{72} & \textbf{66} & \textbf{69} & \textbf{68} & \textbf{73} \\ \hline
\multirow{4}{*}{2:4} & SparseGPT & 62 & 60 & 69 & 73 & 60 & 64 & 67 & 70 \\
 & SDS\tiny{-SparseGPT} & \textbf{65} & \textbf{65} & \textbf{72} & \textbf{76} & \textbf{61} & \textbf{65} & \textbf{70} & \textbf{74} \\ \cline{2-10} 
 & Wanda & 63 & 59 & 69 & 66 & 55 & 67 & 65 & 69 \\
 & SDS\tiny{-Wanda} & \textbf{64} & \textbf{61} & \textbf{69} & \textbf{72} & \textbf{59} & \textbf{69} & \textbf{68} & \textbf{72} \\ \hline
\multirow{4}{*}{4:8} & SparseGPT & 63 & 63 & 71 & 77 & 59 & 65 & 68 & \textbf{71} \\
 & SDS\tiny{-SparseGPT} & \textbf{65} & \textbf{65} & \textbf{74} & \textbf{78} & \textbf{62} & \textbf{65} & \textbf{72} & \textbf{71} \\ \cline{2-10} 
 & Wanda & 60 & 63 & \textbf{72} & 73 & 61 & 66 & 68 & 74 \\
 & SDS\tiny{-Wanda} & \textbf{64} & \textbf{64} & 69 & \textbf{73} & \textbf{61} & \textbf{67} & \textbf{69} & \textbf{77} \\ \hline
\end{tabular}}
\end{table}

\begin{table}[t]
\centering
\caption{Zero-shot performance (accuracy \%) on Lambada}
\label{tab:lambada}
\setlength{\tabcolsep}{0.9mm}{
\renewcommand{\arraystretch}{1.2} 
\begin{tabular}{cccccccccc}
\hline
Sparsity & Method & opt-125m & opt-350m & opt-1.3b & opt-2.7b & gpt2-S & gpt2-M & gpt2-L & gpt2-XL \\ \hline
0 & - & 39.16 & 46.67 & 58.80 & 64.82 & 33.88 & 45.92 & 50.71 & 54.74 \\ \hline
\multirow{4}{*}{50\%} & SparseGPT & 34.81 & 43.51 & 50.32 & \textbf{63.65} & 25.17 & \textbf{46.38} & \textbf{45.37} & 53.29 \\
 & SDS\tiny{-SparseGPT} & \textbf{39.96} & \textbf{48.55} & \textbf{51.68} & \textbf{65.22} & \textbf{32.82} & \textbf{46.11} & \textbf{45.31} & \textbf{53.80} \\ \cline{2-10} 
 & Wanda & 32.54 & 28.97 & \textbf{47.78} & 62.18 & 24.74 & 40.68 & 44.61 & 49.93 \\
 & SDS\tiny{-Wanda} & \textbf{35.40} & \textbf{44.28} & \textbf{55.27} & \textbf{67.61} & \textbf{30.39} & \textbf{45.39} & \textbf{49.10} & \textbf{54.22} \\ \hline
\multirow{4}{*}{2:4} & SparseGPT & 27.55 & 32.33 & 41.57 & 56.10 & 21.89 & 35.69 & 34.02 & 42.85 \\
 & SDS\tiny{-SparseGPT} & \textbf{31.57} & \textbf{37.14} & \textbf{48.53} & \textbf{57.60} & \textbf{27.61} & \textbf{37.55} & \textbf{36.81} & \textbf{45.24} \\ \cline{2-10} 
 & Wanda & 16.26 & 9.61 & 36.77 & 46.81 & 6.42 & 20.32 & 25.54 & 34.93 \\
 & SDS\tiny{-Wanda} & \textbf{22.12} & \textbf{18.03} & \textbf{46.24} & \textbf{60.66} & \textbf{23.93} & \textbf{37.76} & \textbf{40.17} & \textbf{44.11} \\ \hline
\multirow{4}{*}{4:8} & SparseGPT & 30.31 & 37.45 & 45.52 & 59.64 & 24.68 & 40.00 & 40.05 & \textbf{48.61} \\
 & SDS\tiny{-SparseGPT} & \textbf{34.43} & \textbf{42.89} & \textbf{51.68} & \textbf{61.09} & \textbf{27.83} & \textbf{42.83} & \textbf{42.27} & \textbf{48.81} \\ \cline{2-10} 
 & Wanda & 24.35 & 16.36 & \textbf{43.26} & 54.90 & 13.45 & 32.27 & 34.45 & 43.39 \\
 & SDS\tiny{-Wanda} & \textbf{26.57} & \textbf{28.24} & \textbf{51.72} & \textbf{65.30} & \textbf{25.52} & \textbf{42.44} & \textbf{43.94} & \textbf{48.88} \\ \hline
\end{tabular}}
\end{table}

\begin{table}[t]
\centering
\caption{Zero-shot performance (accuracy \%) on OpenBookQA}
\label{tab:openbookqa}
\setlength{\tabcolsep}{0.9mm}{
\renewcommand{\arraystretch}{1.2} 
\begin{tabular}{cccccccccc}
\hline
Sparsity & Method & opt-125m & opt-350m & opt-1.3b & opt-2.7b & gpt2-S & gpt2-M & gpt2-L & gpt2-XL \\ \hline
0 & - & 28.0 & 28.0 & 33.0 & 35.2 & 27.2 & 30.2 & 31.2 & 32.0 \\ \hline
\multirow{4}{*}{50\%} & SparseGPT & 27.0 & 27.6 & 28.6 & \textbf{33.8} & 25.2 & \textbf{28.80} & \textbf{29} & \textbf{31.2} \\
 & SDS\tiny{-SparseGPT} & \textbf{28.2} & \textbf{29.0} & \textbf{32.0} & \textbf{35.2} & \textbf{26.2} & \textbf{29.2} & \textbf{30.0} & \textbf{31.2} \\ \cline{2-10} 
 & Wanda & 26.2 & \textbf{26.2} & \textbf{30.8} & 32.0 & 24.2 & \textbf{29.0} & 28.4 & 30.8 \\
 & SDS\tiny{-Wanda} & \textbf{27.4} & \textbf{26.2} & \textbf{31.4} & \textbf{32.8} & \textbf{24.8} & \textbf{28.8} & \textbf{29.0} & \textbf{31.0} \\ \hline
\multirow{4}{*}{2:4} & SparseGPT & 25.8 & 25.2 & 29.0 & 32.0 & 25.2 & \textbf{28.8} & 26.8 & \textbf{29.8} \\
 & SDS\tiny{-SparseGPT} & \textbf{27.7} & \textbf{28.0} & \textbf{30.0} & \textbf{33.0} & \textbf{25.2} & \textbf{28.4} & \textbf{29.0} & \textbf{29.4} \\ \cline{2-10} 
 & Wanda & 25.8 & \textbf{27.0} & 29.2 & 32.0 & 23.8 & 26.0 & 26.4 & 27.8 \\
 & SDS\tiny{-Wanda} & \textbf{26.2} & \textbf{26.8} & \textbf{29.2} & \textbf{32.0} & \textbf{25.8} & \textbf{27.2} & \textbf{27.2} & \textbf{29.0} \\ \hline
\multirow{4}{*}{4:8} & SparseGPT & 25.8 & 24.8 & 30.0 & 33.2 & 25.0 & 26.2 & \textbf{29.0} & \textbf{30.6} \\
 & SDS\tiny{-SparseGPT} & \textbf{26.4} & \textbf{27.4} & \textbf{31.2} & \textbf{35.0} & \textbf{25.8} & \textbf{28.8} & \textbf{27.2} & \textbf{30.8} \\ \cline{2-10} 
 & Wanda & 24.2 & 24.4 & \textbf{30.6} & 32.0 & 25.0 & \textbf{26.8} & 28.0 & 29.6 \\
 & SDS\tiny{-Wanda} & \textbf{25.4} & \textbf{25.8} & \textbf{29.2} & \textbf{32.0} & \textbf{25.0} & \textbf{26.6} & \textbf{30.6} & \textbf{30.8} \\ \hline
\end{tabular}}
\end{table}

\begin{table}[t]
\centering
\caption{Zero-shot performance (accuracy \%) on PIQA}
\label{tab:piqa}
\setlength{\tabcolsep}{0.9mm}{
\renewcommand{\arraystretch}{1.2} 
\begin{tabular}{cccccccccc}
\hline
Sparsity & Method & opt-125m & opt-350m & opt-1.3b & opt-2.7b & gpt2-S & gpt2-M & gpt2-L & gpt2-XL \\ \hline
0 & - & 62.02 & 64.63 & 72.47 & 74.81 & 62.51 & 66.38 & 69.21 & 70.51 \\ \hline
\multirow{4}{*}{50\%} & SparseGPT & 59.79 & 62.19 & 67.68 & \textbf{72.42} & 60.72 & \textbf{65.18} & \textbf{66.54} & \textbf{68.72} \\
 & SDS\tiny{-SparseGPT} & \textbf{60.61} & \textbf{62.84} & \textbf{70.46} & \textbf{73.23} & \textbf{60.77} & \textbf{65.61} & \textbf{67.14} & \textbf{69.31} \\ \cline{2-10} 
 & Wanda & 59.58 & \textbf{60.83} & \textbf{68.99} & 71.87 & 59.52 & \textbf{63.28} & 66.76 & 67.57 \\
 & SDS\tiny{-Wanda} & \textbf{60.55} & \textbf{61.04} & \textbf{69.42} & \textbf{72.31} & \textbf{60.99} & \textbf{64.53} & \textbf{67.08} & \textbf{70.02} \\ \hline
\multirow{4}{*}{2:4} & SparseGPT & 57.24 & 60.12 & 65.02 & 70.18 & 58.43 & \textbf{62.08} & 64.58 & \textbf{66.81} \\
 & SDS\tiny{-SparseGPT} & \textbf{59.85} & \textbf{60.39} & \textbf{68.44} & \textbf{71.33} & \textbf{60.01} & \textbf{62.79} & \textbf{66.00} & \textbf{68.39} \\ \cline{2-10} 
 & Wanda & 58.27 & \textbf{58.11} & 65.67 & 68.88 & 56.31 & 58.43 & 61.53 & 65.40 \\
 & SDS\tiny{-Wanda} & \textbf{58.41} & \textbf{59.58} & \textbf{66.38} & \textbf{70.13} & \textbf{59.25} & \textbf{62.51} & \textbf{65.34} & \textbf{67.46} \\ \hline
\multirow{4}{*}{4:8} & SparseGPT & 59.03 & 62.13 & 66.97 & 71.82 & 58.41 & 63.06 & \textbf{65.40} & \textbf{66.97} \\
 & SDS\tiny{-SparseGPT} & \textbf{60.34} & \textbf{62.46} & \textbf{69.59} & \textbf{72.91} & \textbf{59.90} & \textbf{63.66} & \textbf{65.94} & \textbf{67.36} \\ \cline{2-10} 
 & Wanda & 58.38 & 60.55 & \textbf{68.01} & 70.78 & 58.16 & \textbf{60.72} & 64.42 & 66.10 \\
 & SDS\tiny{-Wanda} & \textbf{60.07} & \textbf{61.64} & \textbf{68.23} & \textbf{71.93} & \textbf{59.90} & \textbf{63.60} & \textbf{66.92} & \textbf{67.95} \\ \hline
\end{tabular}}
\end{table}

\begin{table}[t]
\centering
\caption{Zero-shot performance (accuracy \%) on RTE}
\label{tab:rte}
\setlength{\tabcolsep}{0.9mm}{
\renewcommand{\arraystretch}{1.2} 
\begin{tabular}{cccccccccc}
\hline
Sparsity & Method & opt-125m & opt-350m & opt-1.3b & opt-2.7b & gpt2-S & gpt2-M & gpt2-L & gpt2-XL \\ \hline
0 & - & 50.18 & 51.99 & 52.35 & 55.23 & 53.07 & 52.71 & 52.71 & 52.35 \\ \hline
\multirow{4}{*}{50\%} & SparseGPT & 46.57 & 53.79 & \textbf{50.54} & \textbf{51.26} & \textbf{53.07} & \textbf{52.35} & \textbf{51.95} & \textbf{50.54} \\
 & SDS\tiny{-SparseGPT} & \textbf{48.74} & \textbf{58.12} & \textbf{50.18} & \textbf{52.71} & \textbf{50.18} & \textbf{52.71} & \textbf{53.89} & \textbf{53.43} \\ \cline{2-10} 
 & Wanda & 49.46 & \textbf{51.60} & \textbf{49.10} & 51.99 & 50.18 & \textbf{52.71} & 53.60 & \textbf{53.43} \\
 & SDS\tiny{-Wanda} & \textbf{51.99} & \textbf{51.99} & \textbf{50.54} & \textbf{52.71} & \textbf{52.35} & \textbf{52.35} & \textbf{53.89} & \textbf{52.35} \\ \hline
\multirow{4}{*}{2:4} & SparseGPT & 53.79 & 53.79 & 50.90 & 52.71 & \textbf{53.07} & \textbf{51.62} & \textbf{52.35} & \textbf{48.47} \\
 & SDS\tiny{-SparseGPT} & \textbf{54.15} & \textbf{54.15} & \textbf{53.79} & \textbf{53.07} & \textbf{52.71} & \textbf{52.35} & \textbf{51.26} & \textbf{57.76} \\ \cline{2-10} 
 & Wanda & 51.26 & \textbf{53.79} & \textbf{50.54} & 52.35 & \textbf{52.71} & 53.79 & 53.79 & 48.38 \\
 & SDS\tiny{-Wanda} & \textbf{52.71} & \textbf{53.79} & \textbf{49.10} & \textbf{52.71} & \textbf{51.62} & \textbf{54.15} & \textbf{53.79} & \textbf{53.79} \\ \hline
\multirow{4}{*}{4:8} & SparseGPT & \textbf{53.07} & 53.07 & 49.82 & 53.07 & 51.62 & 51.89 & \textbf{51.99} & \textbf{53.79} \\
 & SDS\tiny{-SparseGPT} & \textbf{51.99} & \textbf{56.68} & \textbf{53.07} & \textbf{53.79} & \textbf{52.71} & \textbf{52.71} & \textbf{52.35} & \textbf{52.35} \\ \cline{2-10} 
 & Wanda & 50.18 & 51.99 & \textbf{50.90} & \textbf{52.35} & 51.26 & \textbf{53.43} & 52.35 & 50.18 \\
 & SDS\tiny{-Wanda} & \textbf{50.18} & \textbf{52.35} & \textbf{51.99} & \textbf{50.18} & \textbf{51.54} & \textbf{53.79} & \textbf{55.23} & \textbf{54.51} \\ \hline
\end{tabular}}
\end{table}

\begin{table}[t]
\centering
\caption{Zero-shot performance (accuracy \%) on StoryCloze}
\label{tab:storycloze}
\setlength{\tabcolsep}{0.9mm}{
\renewcommand{\arraystretch}{1.2} 
\begin{tabular}{cccccccccc}
\hline
Sparsity & Method & opt-125m & opt-350m & opt-1.3b & opt-2.7b & gpt2-S & gpt2-M & gpt2-L & gpt2-XL \\ \hline
0 & - & 60.03 & 63.27 & 70.78 & 71.74 & 58.18 & 62.70 & 64.42 & 67.85 \\ \hline
\multirow{4}{*}{50\%} & SparseGPT & 58.24 & 60.03 & \textbf{66.71} & \textbf{70.72} & \textbf{55.76} & \textbf{60.85} & \textbf{62.95} & \textbf{66.33} \\
 & SDS\tiny{-SparseGPT} & \textbf{59.13} & \textbf{61.68} & \textbf{68.75} & \textbf{70.72} & \textbf{55.76} & \textbf{61.17} & \textbf{63.40} & \textbf{66.71} \\ \cline{2-10} 
 & Wanda & 57.67 & \textbf{58.18} & \textbf{66.71} & 69.70 & \textbf{55.76} & \textbf{59.26} & 61.17 & \textbf{64.16} \\
 & SDS\tiny{-Wanda} & \textbf{58.18} & \textbf{58.31} & \textbf{67.66} & \textbf{70.02} & \textbf{56.97} & \textbf{60.98} & \textbf{62.76} & \textbf{65.50} \\ \hline
\multirow{4}{*}{2:4} & SparseGPT & 55.38 & 56.33 & 63.84 & 67.66 & \textbf{56.27} & \textbf{57.86} & \textbf{60.79} & \textbf{63.46} \\
 & SDS\tiny{-SparseGPT} & \textbf{57.42} & \textbf{57.73} & \textbf{66.96} & \textbf{69.26} & \textbf{55.82} & \textbf{58.05} & \textbf{61.11} & \textbf{63.72} \\ \cline{2-10} 
 & Wanda & \textbf{54.87} & \textbf{54.36} & \textbf{64.23} & 66.58 & \textbf{53.15} & 54.87 & 57.80 & 60.28 \\
 & SDS\tiny{-Wanda} & \textbf{54.81} & \textbf{55.06} & \textbf{65.75} & \textbf{67.54} & \textbf{54.74} & \textbf{57.22} & \textbf{61.23} & \textbf{62.83} \\ \hline
\multirow{4}{*}{4:8} & SparseGPT & \textbf{56.20} & 57.80 & 64.74 & 69.19 & 54.91 & 59.77 & \textbf{61.74} & \textbf{65.25} \\
 & SDS\tiny{-SparseGPT} & \textbf{58.24} & \textbf{60.09} & \textbf{68.49} & \textbf{69.57} & \textbf{56.46} & \textbf{60.28} & \textbf{61.81} & \textbf{65.93} \\ \cline{2-10} 
 & Wanda & 55.70 & 55.95 & \textbf{64.23} & \textbf{67.99} & 54.11 & \textbf{55.57} & 60.03 & 61.87 \\
 & SDS\tiny{-Wanda} & \textbf{56.21} & \textbf{56.40} & \textbf{66.01} & \textbf{68.43} & \textbf{55.70} & \textbf{59.64} & \textbf{62.38} & \textbf{64.86} \\ \hline
\end{tabular}}
\end{table}

\begin{table}[t]
\centering
\caption{Zero-shot performance (accuracy \%) on Winogrande}
\label{tab:winogrande}
\setlength{\tabcolsep}{0.9mm}{
\renewcommand{\arraystretch}{1.2} 
\begin{tabular}{cccccccccc}
\hline
Sparsity & Method & opt-125m & opt-350m & opt-1.3b & opt-2.7b & gpt2-S & gpt2-M & gpt2-L & gpt2-XL \\ \hline
0 & - & 50.36 & 52.25 & 59.43 & 60.85 & 51.62 & 53.12 & 55.33 & 58.33 \\ \hline
\multirow{4}{*}{50\%} & SparseGPT & \textbf{51.54} & 51.22 & \textbf{56.35} & \textbf{58.12} & \textbf{50.99} & \textbf{52.17} & \textbf{53.51} & \textbf{56.43} \\
 & SDS\tiny{-SparseGPT} & \textbf{50.99} & \textbf{52.41} & \textbf{57.85} & \textbf{60.38} & \textbf{51.85} & \textbf{52.49} & \textbf{53.67} & \textbf{56.55} \\ \cline{2-10} 
 & Wanda & \textbf{51.78} & \textbf{51.54} & \textbf{56.91} & 57.77 & \textbf{51.07} & \textbf{51.14} & 53.04 & \textbf{54.30} \\
 & SDS\tiny{-Wanda} & \textbf{50.91} & \textbf{51.98} & \textbf{57.77} & \textbf{58.96} & \textbf{51.85} & \textbf{52.33} & \textbf{54.30} & \textbf{55.41} \\ \hline
\multirow{4}{*}{2:4} & SparseGPT & 51.14 & 50.59 & 55.64 & 57.70 & \textbf{50.43} & \textbf{51.14} & \textbf{50.43} & \textbf{53.59} \\
 & SDS\tiny{-SparseGPT} & \textbf{51.46} & \textbf{51.07} & \textbf{56.99} & \textbf{59.43} & \textbf{50.99} & \textbf{50.43} & \textbf{52.96} & \textbf{55.25} \\ \cline{2-10} 
 & Wanda & \textbf{50.36} & \textbf{51.54} & \textbf{54.62} & 55.96 & \textbf{48.86} & 51.22 & 52.41 & \textbf{53.83} \\
 & SDS\tiny{-Wanda} & \textbf{51.38} & \textbf{52.55} & \textbf{55.41} & \textbf{57.85} & \textbf{51.54} & \textbf{52.55} & \textbf{53.67} & \textbf{53.59} \\ \hline
\multirow{4}{*}{4:8} & SparseGPT & \textbf{50.59} & 50.67 & 56.51 & 57.77 & 50.59 & 51.38 & \textbf{51.54} & \textbf{54.14} \\
 & SDS\tiny{-SparseGPT} & \textbf{51.30} & \textbf{51.22} & \textbf{57.38} & \textbf{59.98} & \textbf{51.30} & \textbf{51.78} & \textbf{52.57} & \textbf{54.70} \\ \cline{2-10} 
 & Wanda & 51.14 & \textbf{52.64} & \textbf{56.27} & \textbf{56.43} & 49.41 & \textbf{49.25} & 50.91 & 54.14 \\
 & SDS\tiny{-Wanda} & \textbf{51.46} & \textbf{51.85} & \textbf{56.67} & \textbf{58.56} & \textbf{50.83} & \textbf{51.22} & \textbf{53.04} & \textbf{54.30} \\ \hline
\end{tabular}}
\end{table}

\ifx\allfiles\undefined